\documentclass[journal]{IEEEtran}
\usepackage{tikz}
\usepackage{authblk}
\usepackage{amssymb}
\usepackage{amsmath}
\usepackage{gensymb}
\usepackage{url}
\usepackage{layouts}
\usepackage{booktabs}
\usepackage{array}
\usepackage{multirow}
\newcommand{\x}{{\bf x}}

\begin{document}
\date{}
\title{Interpretable Deep Neural Networks\\ for Single-Trial EEG Classification}
\author[1]{Irene Sturm}
\author[2]{Sebastian Bach}
\author[2]{Wojciech Samek}
\author[1,3]{\mbox{Klaus-Robert M{\"u}ller}}
\affil[1]{Machine Learning Group, Berlin Institute of Technology, Berlin, Germany}
\affil[2]{Machine Learning Group, Fraunhofer Heinrich Hertz Institute, Berlin, Germany}
\affil[3]{Department of Brain and Cognitive Engineering, Korea University, Seoul, Korea}
\maketitle 
\begin{abstract}
\noindent {\it Background}: In cognitive neuroscience the potential of Deep Neural Networks (DNNs) for solving complex classification tasks is yet to be fully exploited. The most limiting factor is that DNNs as notorious `black boxes' do not provide insight into neurophysiological phenomena underlying a decision. Layer-wise Relevance Propagation (LRP) has been introduced as a novel method to explain individual network decisions.\\ 
{\it New Method}: We propose the application of DNNs with LRP for the first time for EEG data analysis. Through LRP the single-trial DNN decisions are transformed into heatmaps indicating each data point's relevance for the outcome of the decision.\\
{\it Results}: DNN achieves classification accuracies comparable to those of CSP-LDA. In subjects with low performance subject-to-subject transfer of trained DNNs can improve the results. The single-trial LRP heatmaps reveal neurophysiologically plausible patterns, resembling CSP-derived scalp maps. Critically, while CSP patterns represent class-wise aggregated information, LRP heatmaps pinpoint neural patterns to single time points in single trials.\\ 
{\it Comparison with Existing Method(s)}: We compare the classification performance of DNNs to that of linear CSP-LDA on two data sets related to motor-imaginery BCI.\\ 
{\it Conclusion}: We have demonstrated that DNN is a powerful non-linear tool for EEG analysis. With LRP a new quality of high-resolution assessment of neural activity can be reached. 
LRP is a potential remedy for the lack of interpretability of DNNs that has limited their utility in neuroscientific applications. The extreme specificity of the LRP-derived heatmaps opens up new avenues for investigating neural activity underlying complex perception or decision-related processes.
\end{abstract}

%%%%%%%%%%%%%%%%%%%%%%%%%%%%%%%%%%%%%%%%%%%%%%%%%%%%%%%%%%%%%%%%%%%%%%%%
\section{Introduction}
\label{sec:intro}
Deep Neural Networks (DNNs) are powerful methods for solving complex classification tasks in fields such as computer vision \cite{DBLP:conf/nips/KrizhevskySH12}, natural language processing \cite{Socher-etal:2013}, video analysis \cite{DBLP:conf/cvpr/LeZYN11} and physics \cite{montavon-njp13}. Although researchers have recently started introducing this promising technology into the domain of cognitive neuroscience \cite{plis2014deep} and Brain-Computer Interfacing (BCI) \cite{yuksel2015neural, yang2015use}, most of the current techniques in these fields are still based on linear methods \cite{parra2005recipes, BlaTomLemKawMue08}. A limiting factor for the applicability of DNN in these fields is the notion of a DNN as a {\it black box}. In the domain of cognitive neuroscience this is a particular drawback because obtaining neurophysiological insights is of utmost importance beyond the classification performance of a system.

Recently, the interpretability aspect of deep neural networks has been addressed by the Layer-wise Relevance Propagation (LRP) \cite{BachPLOS15} method. LRP explains individual classification decisions of a DNN by decomposing its output in terms of input variables. It is a principled method which has close relation to Taylor decomposition \cite{MonArXiv15} and is applicable to arbitrary DNN architectures. From a practitioners perspective LRP adds a new dimension to the application of DNNs (e.g., in computer vision \cite{BacCVPR16, SamArxiv15b}) by making the prediction transparent. Within the scope of cognitive neuroscience this means that DNN with LRP, may provide not only a highly effective (non-linear) classification technique that is suitable for complex high-dimensional data, but also yield detailed single-trial accounts of the distribution of decision-relevant information, a feature that is lacking in commonly applied DNN techniques and also in other state-of-the art methods (such as those discussed below).

Here we propose using DNN with LRP for the first time for EEG analysis. For that we train a DNN to solve a classification task related to motor-imaginery BCI. On two example data sets we compare the classification performance of DNN to that of CSP-LDA, a standard technique \cite{BlaTomLemKawMue08}. We then apply LRP to produce {\it heatmaps} that indicate the relevance of each data point of a spatio-temporal EEG epoch for the classifier's decision in single trial. We present several examples of such heatmaps and demonstrate their neurophysiological plausibility. Critically, we point out that the spatio-temporal heatmaps represent a new quality of explanatory resolution that allows to explain why the classifier reaches a certain decision in a single instance.  Note that such information can not be derived from CSP-LDA.
Finally, we provide a range of future applications of this technique in neuroscience. We discuss why equipping the extremely powerful non-linear technology of DNN with the diagnostic power of LRP may contribute to extending the scope of DNN techniques.

%%%%%%%%%%%%%%%%%%%%%%%%%%%%%%%%%%%%%%%%%%%%%%%%%%%%%%%%%%%%%%%%%%%%%%%%
\section{A Deep Neural Network for EEG Classification}
\subsection{Model Details}

The network applied here consists of two linear sum-pooling layers with bias-inputs, followed by an activation or normalization step each. The first linear layer accepts an input of the dimensionality 301 time points $\times$ 118 channels EEG features (301 time point $\times$ 58 channels for subjects od-obx) as a 33518 (od-obx: 17458) dimensional input vector and produces a 500-dimensional tanh-activated output vector. The next layer reduces the 500-dimensional space to a 2-dimensional output space followed by a softmax layer for activation in order to produce output probabilities for each class. The network was trained using a standard error back-propagation algorithm using batches of 5 randomly drawn training samples. The above prediction accuracy was achieved after terminating the training procedure after 3000 iterations \cite{LeCun.2012}.
%%%%
\subsection{Interpretability}
The DNN assigns a classification score $f(\x)$ to every input data sample $\x = [x_1 \ldots x_N]$ at prediction time.
Layer-wise Relevance Propagation decomposes the classifier output $f(\x)$ in terms of relevances $r_i$ attributing to each input component $x_i$ its \emph{share} with which it contributes to the classification decision
\begin{align}
f(\x) = \sum_i r_i
\label{eq:conservation2}
\end{align}

These relevance values are backpropaged from the network output to the input layer using a local redistribution rule
\begin{eqnarray}
r_i^{(l)} = \sum_j \frac{z_{ij}}{\sum_{i'} z_{i'j}} r_j^{(l+1)} \quad \rm{with}\quad z_{ij} = x_i^{(l)} w_{ij}^{(l,l+1)}
\label{eq:LRPnaive}
\end{eqnarray}
where $j$ indexes a neuron at a particular layer $l+1$, where $\sum_i$ runs over all lower-layer neurons connected to neuron $j$, and where $w_{ij}^{(l,l+1)}$ are parameters specific to pairs of adjacent neurons and learned from the data. This redistribution rule has been showed to fulfill the layer-wise conservation property \cite{BachPLOS15} and to be closely related to a deep variant of Taylor decomposition \cite{MonArXiv15}. 

%%%%%%%%%%%%%%%%%%%%%%%%%%%%%%%%%%%%%%%%%%%%%%%%%%%%%%%%%%%%%%%%%%%%%%%%
\section{Evaluation}
\label{sec:evaluation}
\subsection{Experimental Setup and Preprocessing}

The application of DNN with LRP on EEG data was demonstrated on dataset IVa from BCI competition III (cued motor imagery data with classes right hand vs. foot from 5 subjects \cite {BlaMueKruWolSchPfuMilSchBir06}) and on a subset of 5 subjects from \cite{BraNeuralEng15} where subjects had to perform left and right hand motor imaginery while dealing with different types of distractions. Here, we only analyzed data obtained in the condition `no distraction', a standard motor imaginery BCI setting. As in the competition, we did not use test data for training for dataset IVa (subjects aa, al, av, aw, ay). For the other data set (subjects od, njy, njk, nko, obx) a leave-on-out cross-validation was performed. For both data sets the potential of DNN for subject-to-subject transfer was evaluated: for each subject a DNN was trained on all available data of the other four subjects and evaluated on its own test data. This was was done in a sequential fashion, so that the network was once initialized and then trained on the data of each of the four subjects successively. The entire process of training and testing was repeated five times for different orders of the four subjects and the classification performance on the test data was averaged. 

All data sets were downsampled to 100Hz and bandpass filtered in the range of 9-13 Hz. The CSP algorithm was performed on a [1000\ 4000] ms epoch after the cue and 3 pairs of spatial filters were selected. On the extracted features a regularized LDA classifier with analytically determined shrinkage parameter \cite{BlaLemTreHauMue10} was trained. For training and evaluating the DNN the envelope of each epoch ([1000 4000] ms after cue) was calculated and an epochwise baseline of [0 300] ms before the cue was subtracted. Each epoch's spatio-temporal features (301 time points $\times$ 118 channels for aa-ay, 301 time point $\times$ 58 channels for subject od-obx) were vectorized into one vector with 33518 (17458) dimensions. 
Relevance maps were calculated for each trial from the two-valued DNN output according to Equation \ref{eq:LRPnaive}. 
 
%%%%
\subsection{Results}

Classification results for the different methods are summarized in Table \ref{table:class}. Overall, classification performance of DNN is lower than that of CSP-LDA. Subjects ay and njy, the subjects with the lowest performance, represent an exception: here DNN effects an increase in classification accuracy. The performance of inter-subject DNN is inferior to that of single-subject DNN in 6/10 subjects. In the remaining four subjects inter-subject DNN effects a substantial increase in classification accuracy.

\begin{figure*}[h]
\centering
\includegraphics[width=\textwidth]{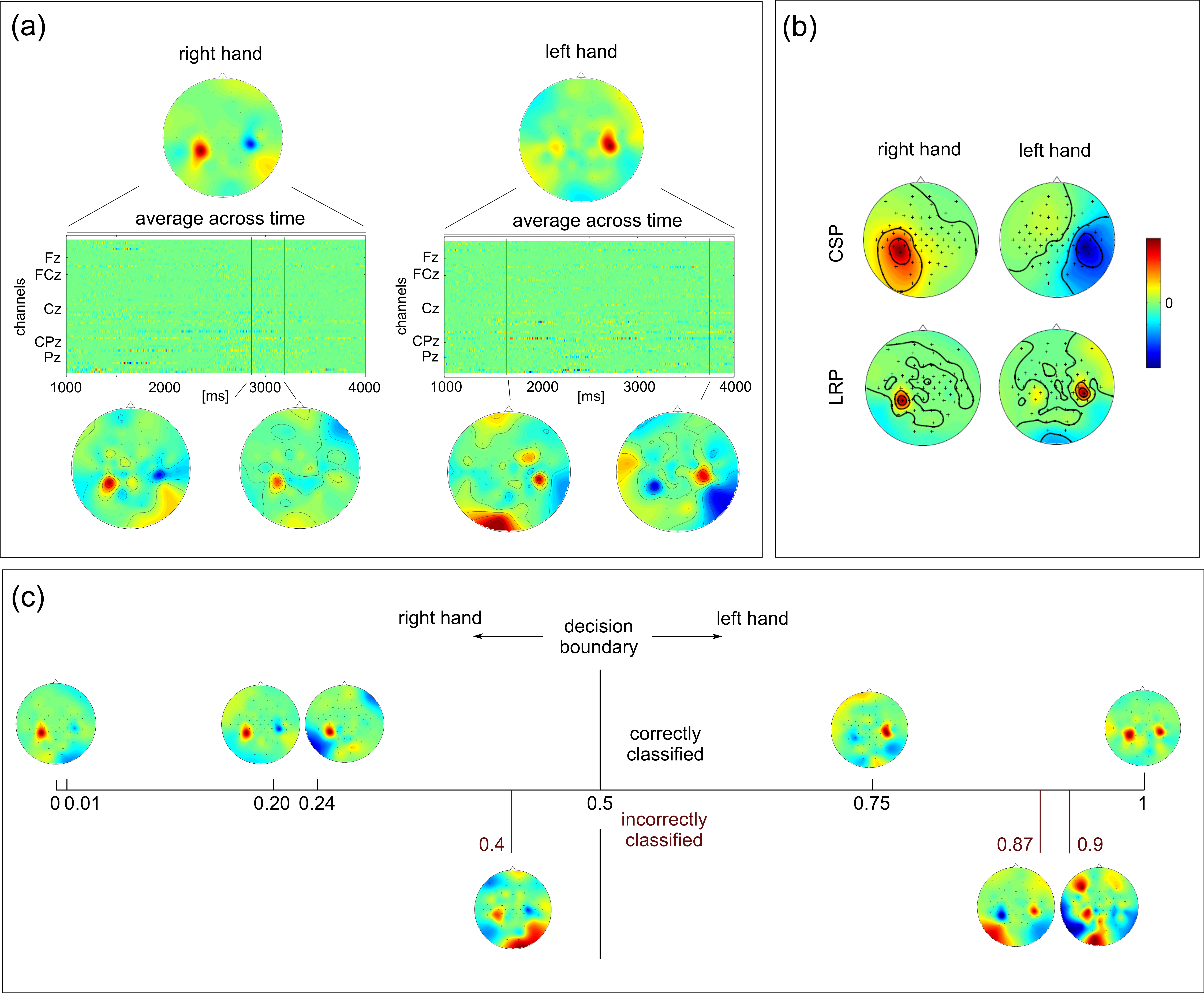}%
\caption{(a) Example of LRP relevance maps for a single trial of each class of subject od. The matrices indicate the relevance of each time point (ordinate) and EEG channel (abscissa). Below the matrix the relevance information for two single time points (indicated by the green line) is plotted as a scalp topography. The scalp plot above the matrices depict the average relevance map across the time window of the entire epoch. (b) CSP patterns (top) and relevance maps (bottom) for subject od. Here, the relevance maps represent the average of all trials of one class, additionally averaged across the time window of the entire epoch. The CSP pattern represents the whole ensemble of samples of one class. (c) Relevance maps and DNN output. Examples of (time-averaged) relevance maps for single trials with different classification outcomes. Values above 0.5 indicate a decision for class `left hand', values below 0.5 for class `right hand'. Values close to the extrema 0 and 1 indicate high confidence of the decision. Correctly classified samples appear above the axis, incorrectly classified samples below. }
\label{fig:1}
\end{figure*}

Fig.~\ref{fig:1} (a) gives an example of relevance maps obtained with LRP for two single trials of subject od. The matrices depict the relevance of each EEG channel at each time point of the epoch. Note that these relevance maps differ from CSP patterns where the \textit{absolute} magnitude of a weight determines its relevance and its sign the polarity. In LRP-derived heatmaps positive and negative values refer to the relevance and non-relevance with respect to the specific decision of the DNN. For instance, in a trial assigned to class `right hand' with high confidence positive values may be understood as speaking {\it for} class `right hand' membership and negative values as speaking {\it against} class `right hand' membership.  For a given time point the relevance information can be plotted as a scalp topography. The example scalp maps at the bottom show typical lateralized motor activation patterns that can be related to a single time point in a single trial. The average of the spatio-temporal relevance matrix across the entire epoch (top) reveals similar scalp patterns. The average of all time-averaged relevance maps of one class (Fig.~\ref{fig:1} (b)) is highly similar in topographical distribution to the patterns of the first pair of CSP filters. Fig.~\ref{fig:1} (c) shows examples of time-averaged relevance maps for a selection of correctly/incorrectly classified trials. In those trials that were classified correctly and with high confidence (classifier output 0 or 1), relevant information is confined to small regions with neurophysiologically highly plausible distribution. In incorrectly or with less confidence classified trials influences outside the sensorimotor areas seem to have influenced the network's decision. These are located in occipital and frontal regions and may indicate the influence of visual activity and of eye movements. 

\begin{table}
\centering
\caption[Classification performance]{Classification accuracies for CSP-LDA, DNN and inter-subject DNN.  Dataset BCI competition III IVa: aa, al, av, aw, ay. Dataset from \cite{BraNeuralEng15}: od, njy, njz, nko, obx}
\begin{tabular}
{c|cc|ccc|}
subject&\multicolumn{2}{ |c |}{number of samples}&\multicolumn{3}{| c |}{class. accuray in \%}\\
\hline
&train&test&CSP/LDA& DNN&inter-subj. DNN\\
\hline
aa&168&112&66&62&56\\
\hline
al&224&56&100&93&83\\
\hline
av&84&196&70&66&64\\
\hline
aw&56&124&99&77&71\\
\hline
ay&28&252&55&60&73\\
\hline
od&71&1&96&94&86\\
\hline
njy&71&1&65&69&62\\
\hline
njz&71&1&93&86&91\\
\hline
nko&71&1&81&57&68\\
\hline
obx&71&1&97&85&100\\
\end{tabular}
\label{table:class}
\end{table}

%%%%%%%%%%%%%%%%%%%%%%%%%%%%%%%%%%%%%%%%%%%%%%%%%%%%%%%%%%%%%%%%%%%%%%%%
\section{Discussion}
\label{sec:discussion}

We have provided the first application of DNN with LRP on EEG data. In terms of classification performance, our relatively simple DNN network does not outperform the benchmark methodology of CSP-LDA. However, we provide some examples that training a network successively on several other subjects is advantageous. For instance, this substantially increased classification accuracy in a subject with particularly low accuracy. This is a first hint that DNN technology may be beneficial for subject-to-subject transfer of learned neural representations, and, ultimately, may advance subject-independent zero training strategies in BCI \cite{FazPopDanBlaMueGr09}.

The most important and novel contribution in this work is the application of LRP. We have demonstrated that LRP produces neurophysiologically highly plausible explanations of how a DNN reaches a decision. More specifically, LRP produced textbook-like motor imaginery patterns in single instants of single trials. These represent accounts of neural activity at an unprecedented level of specificity and detail. In contrast, CSP-LDA (and also other methods) only allow to examine discriminating information at the level of the whole ensemble of samples of one class. In a direct application in the BCI context LRP helped to diagnose influences that led to low-confidence or erroneous decisions of the network.   

Outside BCI DNN with LRP may add a new dimension of explanation in any setting where detailed single-trial information is valued. In clinical applications it may represent a sensitive tool for neurophysiological interpretation of anomalies or differences between populations. Here, the opportunity to integrate prior knowledge about clinical populations through inter-subject DNN analyses may be a further advantage. In contexts where the trial-to-trial variability of EEG is not viewed as a notorious obstacle for analysis, but as a source of information, LRP can contribute high-resolving spatio-temporal representations of underlying neurophysiological phenomena. In particular, this might be interesting for linking brain indices to single instances of behavioral measures \cite{Delorme.2014}, for understanding subtle aspects of complex perceptual processes, such as perception of video or audio quality \cite{PorTreBlaAntSchMoeCurMue13,AcqBosPorCurMueWieBla15}, and of dynamic cognitive processes, such as decision making \cite{Tzovara.2012}.  
Finally, a trained network produces relevance maps for any (even artificially generated) DNN decision. This means that LRP can derive a representation of what a network has learned, e.g., by performing LRP on a `ideal' specimen of a given class or even by systematically exploring the space of possible decisions. This might be an interesting alternative to network visualization techniques \cite{Yosinski.2015}. 

%%%%%%%%%%%%%%%%%%%%%%%%%%%%%%%%%%%%%%%%%%%%%%%%%%%%%%%%%%%%%%%%%%%%%%%%
\section{Conclusion}
In summary, we have provided a showcase of how LRP can add an explanatory layer to the highly effective technique of DNN in the EEG/BCI domain. Our results show that LRP provides highly detailed accounts of relevant information in high-dimensional EEG data that may be useful in analysis scenarios where single trials need to be considered individually. 

\vspace*{0.5cm}
\noindent{\bf Acknowledgement}\\
This work was supported by the Brain Korea 21 Plus Program and by the Deutsche Forschungsgemeinschaft (DFG). This publication only reflects the authors views. Funding agencies are not liable for any use that may be made of the information contained herein. Correspondence to WS and KRM.

%\section*{References}
{\small
\bibliographystyle{elsarticle-num} 
\bibliography{ida,lrp,bibliography}
}
\end{document}